# Dimension Reduction in Singularly Perturbed Continuous-Time Bayesian Networks

**Nir Friedman**
School of Computer Science & Engineering
The Hebrew University
nir@cs.huji.ac.il

**Raz Kupferman**
Institute of Mathematics
The Hebrew University
raz@math.huji.ac.il

## Abstract

*Continuous-time Bayesian networks* (CTBNs) are graphical representations of *multi-component continuous-time Markov processes* as directed graphs. The edges in the network represent direct influences among components. The joint rate matrix of the multi-component process is specified by means of conditional rate matrices for each component separately. This paper addresses the situation where some of the components evolve on a time scale that is much shorter compared to the time scale of the other components. We prove that in the limit where the separation of scales is infinite, the Markov process converges (in distribution, or weakly) to a reduced, or effective Markov process that only involves the slow components. We also demonstrate that for a reasonable separation of scales (an order of magnitude) the reduced process is a good approximation of the marginal process over the slow components. We provide a simple procedure for building a reduced CTBN for this effective process, with conditional rate matrices that can be directly calculated from the original CTBN, and discuss the implications for approximate reasoning in large systems.

## 1 Introduction

Continuous-time Markov processes are a framework used in the modeling of a huge range of stochastic dynamical systems, e.g., chemical kinetics, population dynamics, stock markets, and many more (Gardiner [2004]). We consider Markov processes that are homogeneous in time and have a finite state space. Such systems are fully determined by the state space $S$, the distribution of the process at the initial time, and a description of the dynamics of the process. These dynamics are specified by a *rate matrix* $\mathbb{Q}$, whose off-diagonal entries $q_{a,b}$ are exponential rate intensities for transitioning between states. Intuitively, we can think of the entry $q_{a,b}$ as the rate parameter of an exponential distribution whose value is the duration of time spent in state $a$ before transitioning to $b$. When more than one transition is possible, the shortest duration determines the next state. Thus, $q_{a,b}^{-1}$ is the expected duration in state $a$ before transitioning to state $b$ (assuming it were the only possible transition), and $(\sum_{b \neq a} q_{a,b})^{-1}$ is the expected duration before transitioning out of state $a$.

In many applications, the state space is of the form of a product space $S = S_1 \times S_1 \times \cdots \times S_M$, where $M$ is the number of *components* (such processes are called multi-component). Even if each of the $S_i$ is of low dimension, the dimension of the state space is exponential in the number of components, which often poses computational difficulties, e.g., in learning applications. *Continuous-time Bayesian networks* (Nodelman et al. [2002, 2003]) are a graphical representation for Markov processes that have extra structure, therefore allowing for more a compact representation with fewer parameters. The first assumption is that transitions only occur in one component at a time. Second, the transition rates associated with each component are assumed to only depend on the state of a collection of "parent components". The CTBN is a directed, possibly cyclic graph whose nodes are the components of the process, and whose edges represent parent-child relations.

It is also often the case that certain components are considerably faster than the others. Mathematically, it is assumed that the (conditional) rates associated with the fast components are larger by a factor of $1/\epsilon$ than the (conditional) rates associated with the other, slow components, with $\epsilon \ll 1$. Systems having such property are said to have a *separation of scales*, or to be *singularly perturbed*. Such situations are ubiquitous, for example, in chemical kinetics, where some reactions may occur much faster than other. In such situations, the fast components tend to reach "local equilibrium", relative to the slow components, and under certain conditions, reduced Markovian dynamics can be derived for a lower dimensional system that only involves the slow components (see van Kampen [1985] for a classical review on dimension reduction in scale separated systems; see Givon et al. [2004] for a recent review).

In this paper we derive the limiting Markov process and show how to reduce a CTBN with fast components into a smaller CTBN that involves only the slow components. We discuss the implications of this result for inference in CTBNs with different time scales.

## 2 Continuous-time Bayesian networks

In this section we briefly review the CTBN model (Nodelman et al. [2002]). Consider an $M$-component Markov process
$$X(t) = (X_1(t), X_2(t), \ldots X_M(t))$$
with state space
$$S = S_1 \times S_2 \times \cdots \times S_M.$$

A notational convention: vectors are denoted by boldface symbols, e.g., $X, a$, and matrices are denoted by blackboard style characters, e.g., $\mathbb{Q}$. The states in $S$ are denoted by vectors of indexes, $a = (a_1, \ldots, a_M)$. The indexes $1 \leq i, j \leq M$ are used to enumerate the components.

The dynamics of a time-homogeneous continuous-time Markov process are fully determined by the *Markov transition function*,
$$p_{a,b}(t) = \Pr(X_{t+s} = b | X_s = a),$$
where time-homogeneity implies that the right hand side does not depend on $s$. Provided that the transition function satisfies certain analytical properties (continuity, and regularity; see Chung [1960]) the dynamics are fully captured by a constant matrix $\mathbb{Q}$—the *rate*, or *intensity matrix*—whose entries $q_{a,b}$ are defined by
$$q_{a,b} = \lim_{h \downarrow 0} \frac{p_{a,b}(h) - \delta_{a,b}}{h},$$
where $\delta_{a,b}$ is a multivariate Kronecker delta (an alternative notation using an indicator function is $\mathbb{1}(a = b)$). The Markov process $X_t$ can also be given a pathwise characterization. Suppose the process starts in a state $a$. After spending a finite amount of time in state $a$ it transitions, at a random time, to a random state $b \neq a$. The transition times to the potential new states are exponentially distributed, with $q_{a,b}$, $a \neq b$, being the exponential rate for transitioning from state $a$ to state $b$. The diagonal elements of $\mathbb{Q}$ satisfy the condition that each row sums up to zero. Suppose, for simplicity, that each component $X_j$ takes values in a $d$-dimensional space. Then, the state space is $d^M$-dimensional, and the $\mathbb{Q}$-matrix involves $d^M(d^M - 1)$ parameters.

The time-dependent probability distribution of the process, $p(t)$, whose entries are defined by
$$p_a(t) = \Pr(X(t) = a), \qquad a \in S,$$
satisfies the so-called *master equation*,
$$\frac{d\boldsymbol{p}}{dt} = \mathbb{Q}^T \boldsymbol{p}. \tag{1}$$

It is important to note that the master equation (1) encompasses all the statistical properties of the Markov process. There is a one-to-one correspondence between the description of a Markov process by means of a master equation, and by means of a "pathwise" characterization (up to stochastic equivalence of the latter; see Gikhman and Skorokhod [1975]).

We are concerned here with processes for which every transition involves a single component. In such case, the most general rate matrix takes the form
$$q_{a,b} = \sum_{i=1}^{M} q_{a,b_i}^i \prod_{j \neq i} \delta_{a_j, b_j}, \tag{2}$$
where the $q_{a,b_i}^i$ are the entries of a *conditional rate matrix* $\mathbb{Q}^i$ for $X_i$ transitioning from $a_i$ to $b_i$ given that the state of the system is $a$. The structure (2) represents the fact that each component undergoes transitions independently from the other components, but at a rate that depends on the current state of the entire system. A $\mathbb{Q}$-matrix of the form (2) requires $Md^M(d-1)$ independent parameters, which may still be a large number.

Further reduction in the number of parameters is obtained if additional structure is incorporated. CTBNs are applicable to situations where each of the conditional rate matrices $\mathbb{Q}^i$ is only influenced by a subset of component. Specifically, a parent-child relation is introduced between ordered pairs of components. To every $1 \leq i \leq M$ we define the (possibly empty) set of indexes
$$\text{Par}(i) = \left\{ 1 \leq j \leq M : X_j \text{ is a parent of } X_i \right\},$$
and the state space associated with the parents of $X_i$,
$$S_{\text{Par}(i)} = \underset{j \in \text{Par}(i)}{\bigtimes} S_j.$$

We then introduce a restriction operator $P_i : S \to S_{\text{Par}(i)}$, which extracts from the state of the system the state of the subsystem that consists of the parents of $X_i$,
$$P_i(a) = (a_{m_1}, a_{m_2}, \ldots, a_{m_i}),$$
where $\text{Par}(i) = \{m_1, m_2, \ldots, m_i\}$.

The conditional rate matrix associated with the $i$-th component only depends on the state of the parent components. To make this dependence explicit, we denote the conditional rate matrices by $\mathbb{Q}^{i|\text{Par}(i)}$ with entries $q_{a_i, b_i | P_i(a)}^{i | \text{Par}(i)}$. Thus,

the joint rate matrix of the whole process assumes the reduced form

$$q_{a,b} = \sum_{i=1}^{M} q_{a_i,b_i|P_i(a)}^{i|\text{Par}(i)} \prod_{j \neq i} \delta_{a_j,b_j}. \qquad (3)$$

Equation (3) is, using the terminology of Nodelman et al. [2002], the "amalgamation" of the $M$ conditional rate matrices. Note the compact representation which is valid for both diagonal and off-diagonal entries. It is also noteworthy that amalgamation is a summation, rather than a product; indeed, independent exponential rates are additive. If, for example, every component has $k$ parents, the rate matrix requires now only $Md^{k+1}(d-1)$ parameters.

The dependency relations between components can be represented graphically as a directed graph, $\mathcal{G}$, in which each node corresponds to a component, and each directed edge defines a parent-child relation. A CTBN consists of such a graph, supplemented with a set of $M$ conditional rate matrices $\mathbb{Q}^{i|\text{Par}(i)}$, and an initial distribution. Formally, we define a CTBN as a tuple

$$\mathcal{C} = \left\langle \mathcal{G}, \{\mathbb{Q}^{i|\text{Par}(i)}\}_{i=1}^{M}, P_0 \right\rangle,$$

where $P_0$ is the initial distribution over $X(0)$.

As stated in Nodelman et al. [2002], the graph structure has two main roles: (i) it provides a data structure to which parameters are associated; (ii) it provides a qualitative description of dependencies among the various components of the system. The graph structure also reveals statistical (possibly conditional) independencies between sets of components. An example of a four-component CTBN is shown in Figure 1.

For later use, we note that if we substitute the structure (3) of the rate matrix into the master equation (1), the latter takes the particular form,

$$\frac{dp_b}{dt} = \sum_{i=1}^{M} \sum_{a_i \in S_i} q_{a_i,b_i|P_i(b)}^{i|\text{Par}(i)} p_{(b_1,\ldots,b_{i-1},a_i,b_{i+1},\ldots,b_M)}. \qquad (4)$$

## 3 Singularly perturbed CTBNs

In many situations, it is possible to partition the $M$ components into two sets: "fast" components and "slow" components. A standard measure for the "speed" of a Markov process is the rate at which it equilibrates, which is commonly taken to be the absolute value of the second largest eigenvalue of the $\mathbb{Q}$ matrix. In the context of CTBNs every component is assigned a conditional rate matrix. Scale separation holds if all the equilibration rates associated with the conditional rate matrices of a subset of components are much faster than all the equilibration rates associated with the conditional rate matrices of the complementary components. Intuitively, this means that every fast component has undergone many transitions during a time interval characteristic of a single transition in the slow components. For the sake of mathematical analysis, we will consider CTBNs that are parametrized by a small parameter $\epsilon$ that represents the ratio of a characteristic time of residence in a state of a fast component, and a characteristic time of residence in a state of a slow component.

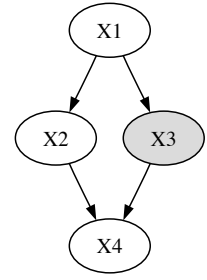

Figure 1: A four-component process, in which $X_1$ is a Markov process (i.e., its rate matrix does not depend on other components), $X_2$ and $X_3$ are influenced by $X_1$, and in turn influence $X_4$. Thus, the construction of a joint rate matrix requires the prescription of a rates matrix $\mathbb{Q}^1$ and conditional rates matrices $\mathbb{Q}^{2|1}$, $\mathbb{Q}^{3|1}$, and $\mathbb{Q}^{4|2,3}$. This structure implies, for example, that $X_1$ and $X_4$ are statistically independent given the *entire trajectories* of $X_2$ and $X_3$.

We now introduce definitions pertinent to the classification of components into fast ones and slow ones: let

$$I_{\text{fast}} = \{1 \leq i \leq M : X_i \text{ is a fast component}\}$$
$$I_{\text{slow}} = \{1 \leq i \leq M : X_i \text{ is a slow component}\}$$

be the sets of indexes of fast and slow components, respectively, and let

$$S_{\text{fast}} = \bigtimes_{i \in I_{\text{fast}}} S_i \qquad \text{and} \qquad S_{\text{slow}} = \bigtimes_{i \in I_{\text{slow}}} S_i$$

be the state spaces associated with the two sets of components. For $a \in S$ we define the restriction operator Fast : $S \to S_{\text{fast}}$,

$$\text{Fast}(a) = (a_{i_1}, a_{i_2}, \ldots, a_{i_m}),$$

with $\{i_1, i_2, \ldots, i_m\} = I_f$. The restriction operator Slow : $S \to S_{\text{slow}}$ is defined similarly. Below we will derive expressions that involves states in $S$, $S_{\text{slow}}$ and $S_{\text{fast}}$. For the sake of readability, we will use the symbols $a, b$ for states in $S$, $\alpha, \beta$ for states in $S_{\text{slow}}$, and $\zeta$ for states in $S_{\text{fast}}$.

We make the following assumptions:

**Assumption 3.1** *The conditional rate matrices associated with fast components can be expressed as $1/\epsilon$ times an $\epsilon$-independent rate matrix, while the conditional rate matrices associated with the slow components do not depend on*

$\epsilon$. Furthermore, for every fixed state of the slow components, the Markov process defined by the conditional rate matrices of the fast components is ergodic over $S_{fast}$.

Note that even if the entire system is ergodic, the subsystem that consists only of the fast components, with the slow components fixed, is not necessarily ergodic. Thus, we need to explicitly require this additional property. This condition is automatically satisfied, for example, if the conditional rate matrices have strictly positive off-diagonal entries. We will denote the (conditional) equilibrium distribution of the fast components, given the state of the slow components, by $\pi^{I_{fast}|I_{slow}}$ with entries $\pi^{I_{fast}|I_{slow}}_{\zeta|\alpha}$ (where, as stated above, $\zeta \in S_{fast}$ and $\alpha \in S_{slow}$).

Graphically, we will mark fast components by nodes with shaded fillings. For example, Figure 1 represents a four-component CTBN in which only $X_3$ is a fast component. Consequently,

$$I_{fast} = \{3\} \qquad I_{slow} = \{1,2,4\}$$
$$S_{fast} = S_3 \qquad S_{slow} = S_1 \times S_2 \times S_4$$
$$\text{Fast}(a) = a_3 \qquad \text{Slow}(a) = (a_1, a_2, a_4),$$

and the conditional rate matrices can be written as $\mathbb{Q}^1$, $\mathbb{Q}^{2|1}$, $\frac{1}{\epsilon}\mathbb{Q}^{3|1}$ and $\mathbb{Q}^{4|2,3}$, where each of the $\mathbb{Q}^{i|\text{Par}(i)}$ is $\epsilon$-independent.

The goal is to study the limiting behavior of the system as $\epsilon \to 0$. Our main theorem, whose proof is sketched in Appendix A, is:

**Theorem 3.1** *Let $X(t)$ be an M-component Markov process satisfying Assumption 3.1. Then as $\epsilon \to 0$, the distribution $p(t)$ converges to a product distribution of the form,*

$$p_a(t) = \pi^{I_{fast}|I_{slow}}_{\text{Fast}(a)|\text{Slow}(a)} \tilde{p}_{\text{Slow}(a)}(t),$$

*where $\tilde{p}$ is the marginal distribution of the slow components. The latter, satisfies the reduced, or effective master equation,*

$$\frac{d}{dt}\tilde{p} = \tilde{\mathbb{Q}}^T \tilde{p}, \qquad (5)$$

*where $\tilde{\mathbb{Q}}$ is a rate matrix over $S_{slow}$. Its entries are given by*

$$\tilde{q}_{\alpha,\beta} = \sum_{i \in I_{slow}} \tilde{q}^i_{\alpha,\beta} \prod_{I_{slow} \ni j \neq i} \delta_{\alpha_j,\beta_j}, \qquad (6)$$

*where $\tilde{\mathbb{Q}}^i$ with entries*

$$\tilde{q}^i_{\alpha,\beta} = \sum_{\zeta \in S_{fast}} \pi^{I_{fast}|I_{slow}}_{\zeta|\alpha} q^{i|\text{Par}(i)}_{\alpha_i,\beta_i|\text{P}_i((\alpha,\zeta))} \qquad (7)$$

*is the effective conditional rate matrix associated with the slow component $X_i$.*

The fact that the marginal distribution over $S_{slow}$ satisfies a master equation means that the limiting behavior of the slow components is Markovian.

*Remark.* When interpreting (7) note that $\alpha$ and $\beta$ are both elements in $S_{slow}$, whereas the summation variables $\zeta$ are elements in $S_{fast}$. The concatenation $(\alpha, \zeta)$ is identified as an element of the full space $S$, and $\text{P}_i((\alpha, \zeta))$ restricts the state $(\alpha, \zeta)$ to only those components that are parents of $X_i$.

The expression (6) for the effective rates can be written in an alternative form. Writing

$$\mathbb{Q} = \frac{1}{\epsilon}\mathbb{Q}^{fast} + \mathbb{Q}^{slow},$$

where $\mathbb{Q}^{fast}$ and $\mathbb{Q}^{slow}$ are $\epsilon$-independent and have entries

$$q^{fast}_{a,b} = \sum_{i \in I_{fast}} q^{i|\text{Par}(i)}_{a_i,b_i|\text{P}_i(a)} \prod_{j \neq i} \delta_{a_j,b_j}$$

$$q^{slow}_{a,b} = \sum_{i \in I_{slow}} q^{i|\text{Par}(i)}_{a_i,b_i|\text{P}_i(a)} \prod_{j \neq i} \delta_{a_j,b_j},$$

the entries of $\tilde{\mathbb{Q}}$ can be written as

$$\tilde{q}_{\alpha,\beta} = \sum_{\zeta \in S_{fast}} \pi^{I_{fast}|I_{slow}}_{\zeta|\alpha} q^{slow}_{(\alpha,\zeta),(\beta,\zeta)}. \qquad (8)$$

In simple words, the reduced rate matrix associated with the limiting dynamics of the slow components is the full rate matrix, averaged over the conditional equilibrium distribution of the fast components. If we denote by $\mathbb{E}^{f|s}$ expectation with respect to the conditional distribution $\pi^{I_{fast}|I_{slow}}$, then (8) takes the more suggestive form,

$$\tilde{\mathbb{Q}} = \mathbb{E}^{f|s}[\mathbb{Q}^{slow}].$$

**Example 3.1** As the simplest illustration of Theorem 3.1, consider a two-component system $X(t) = (X_1(t), X_2(t))$ with rate matrix of the form

$$\mathbb{Q} = \frac{1}{\epsilon}\mathbb{Q}^1 + \mathbb{Q}^{2|1}.$$

That is, $X_1$ is a fast component and it influences the slow component $X_2$. The equilibrium distribution of $X_1$ is denoted by $\pi^1$. Theorem 3.1 asserts that as $\epsilon \to 0$, $X_2(t)$ converges in a weak sense (i.e., in distribution) to a Markov process with effective rate $\tilde{\mathbb{Q}}$, whose entries are given by

$$\tilde{q}_{\alpha_2,\beta_2} = \tilde{q}^2_{\alpha_2,\beta_2} = \sum_{\zeta_1 \in S_1} \pi^1_{\zeta_1} q^{2|1}_{\alpha_2,\beta_2|\zeta_1}.$$

The next section addresses the systematic derivation of reduced CTBNs.

# 4 Dimension reduction of CTBNs

## 4.1 Segregated fast components

We start by considering CTBNs in which fast components are segregated: there are no parent-child relations between two fast components. In such case the conditional equilibrium distribution of the fast components factors into a product distribution on the form

$$\pi^{I_{\text{fast}}|I_{\text{slow}}}_{\zeta|\alpha} = \prod_{i \in I_{\text{fast}}} \pi^{i|\text{Par}(i)}_{\zeta_i|P_i(\alpha)}, \qquad (9)$$

where, with a slight abuse of notations, $P_i(\alpha)$ extracts from the vector $\alpha \in S_{\text{slow}}$ those components that belong to $\text{Par}(i)$. That is, the marginal equilibrium distribution of each fast components depends only on the state of its parent components, which by assumption are all slow components.

Substituting the factorization (9) into the effective rate matrix (7), we obtain

$$\tilde{q}^i_{\alpha,\beta} = \sum_{\zeta} \left[ \prod_k \pi^{k|\text{Par}(k)}_{\zeta_k|P_k(\alpha)} \right] q^{i|\text{Par}(i)}_{\alpha_i,\beta_i|(P_i(\alpha),\zeta)},$$

where the product is over $k \in I_{\text{fast}} \cap \text{Par}(i)$ (i.e., fast parents of $X_i$) and the sum is over $\zeta$ in the corresponding state space $S_{\text{fast}} \cap S_{\text{Par}(i)}$. Note that $\tilde{\mathbb{Q}}^i$ is only conditional on those components that are either slow parents of $X_i$, or (non-exclusively) slow parents of fast parents of $X_i$.

*Remark.* In such cases where the fast dynamics can be factored into independent components, there is no necessity for all fast components to evolve on the same fast scale. The results remain unchanged if each fast component has its own time scale $\epsilon_i$, as long as $\epsilon_i \to 0$ for all $i \in I_{\text{fast}}$.

**Example 4.1** Consider the three-component system depicted in Figure 2 (left), which consists of a chain of three components, the one in the middle being fast. The dynamics are defined by the conditional rate matrices $\mathbb{Q}^1$, $\mathbb{Q}^{2|1}$ and $\mathbb{Q}^{3|2}$. Let $\pi^{2|1}$ be the equilibrium distribution of $X_2$ given a fixed state of $X_1$. Theorem 3.1 implies that as $\epsilon \to 0$, the joint distribution $\tilde{p}$ of the slow components $X_1, X_3$ tends to the solution of a master equation which corresponds to the reduced two-state CTBN shown in Figure 2 (right). The effective rate matrix $\tilde{\mathbb{Q}}$ is determined by the conditional rate matrices $\tilde{\mathbb{Q}}^1$ and $\tilde{\mathbb{Q}}^{3|1}$ given by

$$\tilde{q}^1_{\alpha_1,\beta_1} = q^1_{\alpha_1,\beta_1}$$
$$\tilde{q}^{3|1}_{\alpha_3,\beta_3|\alpha_1} = \sum_{\zeta_2 \in S_2} \pi^{2|1}_{\zeta_2|\alpha_1} q^{3|2}_{\alpha_3,\beta_3|\zeta_2}.$$

**Example 4.2** Consider the CTBN shown in Figure 3 (left). The dynamics are defined by the conditional rate matrices

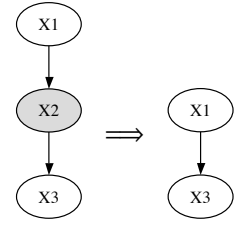

Figure 2: Left: a three-component CTBN with one fast component. Right: the reduced two-component CTBN in the limit $\epsilon \to 0$.

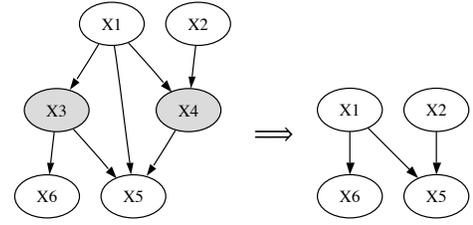

Figure 3: Left: a six-component CTBN with two fast components. Right: the reduced four-component CTBN in the $\epsilon \to 0$ limit.

$\mathbb{Q}^1$, $\mathbb{Q}^2$, $\mathbb{Q}^{3|1}$, $\mathbb{Q}^{4|1,2}$, $\mathbb{Q}^{5|1,3,4}$ and $\mathbb{Q}^{6|3}$. The components $X_3, X_4$ are assumed to be fast, and have conditional equilibrium distributions $\pi^{3|1}$ and $\pi^{4|1,2}$, respectively. As $\epsilon \to 0$, the slow components weakly converge to a four-component Markov process on $S_{\text{slow}}$ defined by the conditional rate matrices $\tilde{\mathbb{Q}}^1$, $\tilde{\mathbb{Q}}^2$, $\tilde{\mathbb{Q}}^{5|1,2}$, and $\tilde{\mathbb{Q}}^{6|1}$ whose entries are given by

$$\tilde{q}^1_{\alpha_1,\beta_1} = q^1_{\alpha_1,\beta_1}$$
$$\tilde{q}^2_{\alpha_2,\beta_2} = q^2_{\alpha_2,\beta_2}$$
$$\tilde{q}^{5|1,2}_{\alpha_5,\beta_5|(\alpha_1,\alpha_2)} = \sum_{\zeta_3 \in S_3} \pi^{3|1}_{\zeta_3|\alpha_1} \sum_{\zeta_4 \in S_4} \pi^{4|1,2}_{\zeta_4|(\alpha_1,\alpha_2)} q^{5|1,3,4}_{\alpha_5,\beta_5|(\alpha_1,\zeta_3,\zeta_4)}$$
$$\tilde{q}^{6|1}_{\alpha_6,\beta_6|\alpha_1} = \sum_{\zeta_3 \in S_3} \pi^{3|1}_{\zeta_3|\alpha_1} q^{6|3}_{\alpha_6,\beta_6|\zeta_3}.$$

Note that although $X_6$ and $X_5$ are statistically dependent, both being descendants of $X_3$, they do not directly influence each other in the reduced CTBN; in general, the elimination of a fast component does not introduce graphical connections between its descendants, in contrast to node elimination in Bayesian networks. This point is highly non-trivial: for finite $\epsilon$ the knowledge of $X_5$ influences the posterior distribution of $X_3$, which in turn affects the evolution of $X_6$. In the limit of extreme scale separation, $X_3$ equilibrates between successive transitions of the slow components, therefore the effective rate matrix of $X_6$ is only affected by the stationary distribution of $X_3$, which, in turn, is only affected by $X_1$.

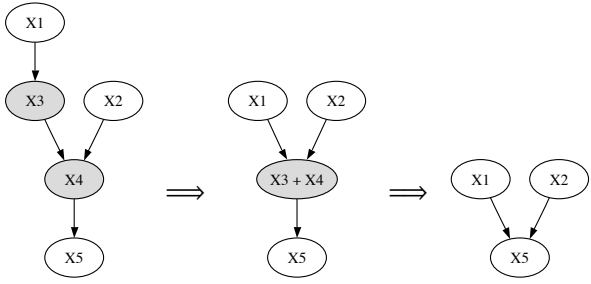
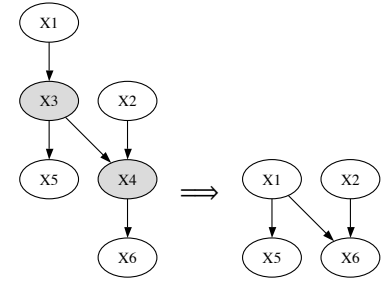

Figure 4: Left: a five-component CTBN with two interacting fast component. Center: equivalent CTBN with the interacting fast components grouped together. Right: the reduced three-component CTBN in the $\epsilon \to 0$ limit.

Figure 5: Left: a six-component CTBN with two interacting fast component. Right: the reduced four-component CTBN in the $\epsilon \to 0$ limit.

### 4.2 Connected fast components

In situations where parent-child relations exist between fast components, the equilibrium distribution on $S_{\text{fast}}$ conditional of $S_{\text{slow}}$ does not factor into a product over fast components. A simple minded solution is to group together fast components that are connected together into a compound component, as illustrated in the following example:

**Example 4.3** Consider the CTBN shown in Figure 4 (left), which is defined by the conditional rate matrices $\mathbb{Q}^1$, $\mathbb{Q}^2$, $\mathbb{Q}^{3|1}$, $\mathbb{Q}^{4|3,2}$ and $\mathbb{Q}^{5|4}$. The component $X_3, X_4$ are assumed to be fast. This situation can be adapted to the framework considered in the previous subsection by grouping together $X_3$ and $X_4$ into a single component with state space $S_3 \times S_4$. This intermediate CTBN is depicted in Figure 4 (center). Let then $\pi^{3,4|1,2}$ denote the equilibrium distribution of the pair $X_3, X_4$ given the pair $X_1, X_2$. As $\epsilon \to 0$, the slow components weakly converge to the Markov processes depicted in the right, with conditional rate matrices $\tilde{\mathbb{Q}}^1$, $\tilde{\mathbb{Q}}^2$, and $\tilde{\mathbb{Q}}^{5|1,2}$, whose entries are given by

$$\tilde{q}^1_{\alpha_1,\beta_1} = q^1_{\alpha_1,\beta_1}$$
$$\tilde{q}^2_{\alpha_2,\beta_2} = q^2_{\alpha_2,\beta_2}$$
$$\tilde{q}^{5|1,2}_{\alpha_5,\beta_5|(\alpha_1,\alpha_2)} = \sum_{\zeta_3 \in S_3} \sum_{\zeta_4 \in S_4} \pi^{3,4|1,2}_{(\zeta_3,\zeta_4)|(\alpha_1,\alpha_2)} q^{5|4}_{\alpha_5,\beta_5|\zeta_4}.$$

This solution, however, misses some of the structure in the reduced process.

**Example 4.4** Consider the CTBN shown in Figure 5 (left), which is defined by the conditional rate matrices $\mathbb{Q}^1$, $\mathbb{Q}^2$, $\mathbb{Q}^{3|1}$, $\mathbb{Q}^{4|3,2}$, $\mathbb{Q}^{5|3}$, and $\mathbb{Q}^{6|4}$. The components $X_3, X_4$ are assumed to be fast. If we group $X_3$ and $X_4$, we again have that $\pi^{3,4|1,2}$ describes the equilibrium distribution of $X_3$ and $X_4$ given $X_1$ and $X_2$. This would imply that $X_5$ depends on both $X_1$ and $X_2$ in the reduced CTBN. However, if we examine the pattern of influence in the original CTBN, the intuition is that there is no path of influence from $X_2$ to $X_5$.

The situation becomes clearer once we realize that we can represent the equilibrium distribution of $X_3$ and $X_4$ as

$$\pi^{3,4|1,2}_{(\zeta_3,\zeta_4)|(\alpha_1,\alpha_2)} = \pi^{3|1}_{\zeta_3|\alpha_1} \pi^{4|1,2,3}_{\zeta_4|(\alpha_1,\alpha_2,\zeta_3)}.$$

To see this, note that $X_4$ does not influence $X_3$. Thus, if we fix the values of $X_1$, the process $X_3$ is Markovian and reaches the same equilibrium distribution regardless of the state of $X_2$. Repeating the same argument as in the previous example, we conclude that as $\epsilon \to 0$, the slow components weakly converge to the Markov processes depicted on the right, with conditional rate matrices $\tilde{\mathbb{Q}}^1$, $\tilde{\mathbb{Q}}^2$, $\tilde{\mathbb{Q}}^{5|1}$, and $\tilde{\mathbb{Q}}^{6|1,2}$, whose entries are given by

$$\tilde{q}^1_{\alpha_1,\beta_1} = q^1_{\alpha_1,\beta_1}$$
$$\tilde{q}^2_{\alpha_2,\beta_2} = q^2_{\alpha_2,\beta_2}$$
$$\tilde{q}^{5|1}_{\alpha_5,\beta_5|\alpha_1} = \sum_{\zeta_3 \in S_3} \pi^{3|1}_{\zeta_3|\alpha_1} q^{5|3}_{\alpha_5,\beta_5|\zeta_3}$$
$$\tilde{q}^{6|1,2}_{\alpha_6,\beta_6|(\alpha_1,\alpha_2)} = \sum_{\zeta_3 \in S_3} \sum_{\zeta_4 \in S_4} \pi^{3,4|1,2}_{(\zeta_3,\zeta_4)|(\alpha_1,\alpha_2)} q^{6|4}_{\alpha_6,\beta_6|\zeta_4}.$$

Note that in the last equation we sum over $\zeta_3$, for the purpose of finding the marginal probability of $X_4$ in $\pi^{3,4|1,2}$.

The point of the last example is that although we cannot factor the equilibrium distribution of the two fast components $X_3$ and $X_4$, the marginal distribution of one variable, $X_3$ in this example, does not depend on the conditional rate matrix of the other variable.

To generalize this line of reasoning, we need to characterize which marginal distributions of the equilibrium distribution can be computed independently of rates of the other components. To analyze such situations, we consider a more general result about marginal distributions in CTBNs.

**Definition 4.1** *Let $\mathcal{C}$ be an M-component CTBN, and let $J$ be a subset of the components $1, \ldots, M$; we denote by*

$$X^J(t) = \{X_i(t) : i \in J\}$$

*the corresponding sub-process. We say that $J$ is* upward closed *if for every $i \in J$, we have that $\text{Par}(i) \subseteq J$. We define the* upward closure $\text{Up}(J)$ *to be the minimal upward closed set that contains $J$.*

**Example 4.5** In Example 4.4 (Figure 5) $J = \{1, 3, 5\}$ is an upward closed subset of components, and

$$\text{Up}(\{4\}) = \{1, 2, 3, 4\}.$$

Suppose we are given a CTBN $\mathcal{C}$. Given an upward closed subset of components, $J$, we can define the *sub-CTBN* spanned by $J$. Formally, we define

$$\mathcal{C}_J = \left\langle \mathcal{G}|_J, \left\{\mathbb{Q}^{i|\text{Par}(i)}\right\}_{i \in J}, P_0|_J \right\rangle,$$

where $\mathcal{G}|_J$ is the sub-graph of $\mathcal{G}$ restricted to the components in $J$, and $P_0|_J$ is the marginal distribution of $P_0$ over $X^J(0)$. The sub-CTBN spanned by $J$, contains the conditional rate matrices from the original CTBN for all the components in $J$. Since $J$ is upward closed, this results in a well defined CTBN, as the parents of every component in $J$ appear in the sub-CTBN.

**Theorem 4.1** *Let $\mathcal{C}$ be an M-component CTBN, and let $J$ be an upward closed subset of components. Then, the marginal distribution over $X^J(t)$ in $\mathcal{C}$ is identical to the distribution over $X^J(t)$ in $\mathcal{C}_J$.*

The proof is straightforward, as we can show that the probability over any trajectory of $X^J$ is the same in both distributions. This follows, for example, if we sum the master equation (4) over all indexes, $b_i, i \notin J$.

**Corollary 4.1** *Let $\mathcal{C}$ be an M-component CTBN, and let $J$ be an upward closed subset of components. Then, the marginal equilibrium distribution over $X^J$ does not depend on the rates associated with the remaining variables.*

Using this result we can return to the question of elimination of fast components. Suppose we have a connected set of fast components. Since they are much faster than the slow components, we can view their behavior as though the slow components are fixed. This implies that the equilibrium distribution over the connected fast components has the behavior of the *conditional-CTBN* defined by their conditional rate matrices, with the slow components fixed. In this conditional CTBN, we can apply Corollary 4.1 and find the set of conditional rate matrices that determine the equilibrium distribution of any particular subset of fast components.

For example, this result implies that in Example 4.4 the equilibrium distribution over $X_3$ in the "fast" conditional CTBN depends on the rate matrix $\mathbb{Q}^{3|1}$ but not on $\mathbb{Q}^{4|2,3}$. As a consequence the child, $X_5$, of $X_3$ depends on $X_1$ in the reduced CTBN, but not on $X_2$.

We now use this intuition to define the reduced CTBN in a precise manner. Consider a CTBN with fast and slow components. Given a set, $J \subseteq I_{\text{fast}}$ of fast components, we define $\text{Up}_f(J)$ to be the upward closure of $J$ in the subgraph that only consists of fast component; $\text{Up}_f(J)$ is the smallest subset of fast components that contains $J$, such that if $i \in J$, then $j \in J$ for all $j \in \text{Par}(i) \cap I_{\text{fast}}$; to shorten the terminology, we will call $\text{Up}_f(J)$ the *fast-upward closure* of $J$. We then define for $J \subseteq I_{\text{fast}}$ the set

$$\text{sPar}(J) = \left\{i \in I_{\text{slow}} : \exists j \in \text{Up}_f(J), i \in \text{Par}(j)\right\}.$$

That is, $\text{sPar}(J)$ are the slow components that are parents of components in $J$, or components in the fast upward closure of $J$. We will call $\text{sPar}(J)$ the set of *last slow ancestors* of $J$.

**Example 4.6** Consider once again Example 4.4 (Figure 5). There,

$$\text{Up}_f(\{4\}) = \{3, 4\}$$

is the upward closure of the subset of components $\{4\}$, in the subgraph that only contains the fast components. Moreover,

$$\text{sPar}(\{4\}) = \{1, 2\},$$

since $X_2$ is a parent of $X_4$ and $X_1$ is a parent of $X_3$, which belongs to fast-upward closure of $\{4\}$.

We now can formally define the procedure of building a reduced CTBN. Assume we are given a CTBN $\mathcal{C}$ with scale separation. We define the reduced CTBN $\tilde{\mathcal{C}}$ as follows:

1. $\tilde{\mathcal{G}}$ is the graph over $I_{\text{slow}}$ such that for each $i \in I_{\text{slow}}$

   $$\widetilde{\text{Par}}(i) = (\text{Par}(i) \cap I_{\text{slow}}) \cup \text{sPar}(\text{Par}(i) \cap I_{\text{fast}})$$

   In other words, the parents of each slow component $X_i$ in the reduced CTBN $\tilde{\mathcal{C}}$ are its slow parents in $\mathcal{C}$ supplemented by the last slow ancestors of its fast parents in $\mathcal{C}$. Consistently with out notations we define $\tilde{P}_i(\alpha)$ to be the restriction of $\alpha \in S_{\text{slow}}$ to those components that belong to $\widetilde{\text{Par}}(i)$.

2. For each $i \in I_{\text{slow}}$ we define the conditional rate matrix $\tilde{\mathbb{Q}}^{i|\widetilde{\text{Par}}(i)}$ with entries

$$\tilde{q}^{i|\widetilde{\text{Par}}(i)}_{\alpha_i,\beta_i|\tilde{P}_i(\alpha)} = \sum_\zeta \pi^{\text{Up}_f(\text{Par}(i) \cap I_{\text{fast}})|\widetilde{\text{Par}}(i)}_{\zeta|\tilde{P}_i(\alpha)} q^{i|\text{Par}(i)}_{\alpha_i,\beta_i|P_i((\alpha,\zeta))},$$

where the summation variable $\zeta$ takes values in the state space spanned by the fast-upper closure of the fast parents of $X_i$. While the last equation may be difficult to parse, it bears a simple interpretation. The conditional rate matrix of the $i$-th component in the reduced CTBN is obtained by averaging over $\mathbb{Q}^{i|\text{Par}(i)}$ with respect to the marginal equilibrium distribution of the fast-upward closure of the fast parents of $X_i$, conditioned by the last slow ancestors of this fast-closure. The effective conditional rate matrix depends, as a result, only on the slow parents of $X_i$ and on the last slow ancestors of its fast parents.

3. $\tilde{P}_0 = P_0|_{I_{\text{slow}}}$.

Our main Theorem 3.1, reformulated in the language of CTBNs, implies:

**Theorem 4.2** *Let $\mathcal{C}$ be an M-component CTBN with conditional rate matrices satisfying Assumption 3.1. Then, as $\epsilon \to 0$, the marginal distribution of the sub-process spanned by the slow components $I_{\text{slow}}$ converges to the distribution induced by the reduced CTBN $\tilde{\mathcal{C}}$.*

## 5 Numerical examples

**Example 5.1** Consider Example 4.1 with state space $S_j = \{0,1\}$, $j = 1,2,3$, and conditional rate matrices

$$\mathbb{Q}^1 = \begin{pmatrix} -1 & 1 \\ 2 & -2 \end{pmatrix}$$

$$\mathbb{Q}^{2|1}_{\cdot|0} = \begin{pmatrix} -2 & 2 \\ 3 & -3 \end{pmatrix} \quad \mathbb{Q}^{2|1}_{\cdot|1} = \begin{pmatrix} -3 & 3 \\ 2 & -2 \end{pmatrix}$$

$$\mathbb{Q}^{3|2}_{\cdot|0} = \begin{pmatrix} -3 & 3 \\ 4 & -4 \end{pmatrix} \quad \mathbb{Q}^{3|2}_{\cdot|1} = \begin{pmatrix} -5 & 5 \\ 6 & -6 \end{pmatrix}$$

The conditional equilibrium distribution $\pi^{2|1}$ is

$$\pi^{2|1}_{\cdot|0} = \frac{1}{5}\begin{pmatrix} 3 \\ 2 \end{pmatrix} \quad \pi^{2|1}_{\cdot|1} = \frac{1}{5}\begin{pmatrix} 2 \\ 3 \end{pmatrix}.$$

As $\epsilon \to 0$ the slow components $X_1, X_3$ weakly converge to a Markov process with effective rate matrices $\tilde{\mathbb{Q}}^1 = \mathbb{Q}^1$, and $\tilde{\mathbb{Q}}^{3|1}$ given by

$$\tilde{\mathbb{Q}}^{3|1}_{\cdot|0} = \pi^{2|1}_{0|0}\mathbb{Q}^{3|2}_{\cdot|0} + \pi^{2|1}_{1|0}\mathbb{Q}^{3|2}_{\cdot|1} = \frac{1}{5}\begin{pmatrix} -19 & 19 \\ 24 & -24 \end{pmatrix}$$

$$\tilde{\mathbb{Q}}^{3|1}_{\cdot|1} = \pi^{2|1}_{0|1}\mathbb{Q}^{3|2}_{\cdot|0} + \pi^{2|1}_{1|1}\mathbb{Q}^{3|2}_{\cdot|1} = \frac{1}{5}\begin{pmatrix} -21 & 21 \\ 26 & -26 \end{pmatrix}.$$

The total effective matrix is

$$\tilde{\mathbb{Q}} = \begin{pmatrix} -4.75 & 1 & 3.75 & 0 \\ 2 & -6.25 & 0 & 4.25 \\ 4.75 & 0 & -5.75 & 1 \\ 0 & 5.25 & 2 & -7.25 \end{pmatrix},$$

with a lexicographic ordering of the states.

To test the accuracy of the procedure we have generated paths of length $T = 50000$ and used standard maximum likelihood to estimate the rate matrix, assuming that the process $(X_1(t), X_3(t))$ is a Markov process. For $\epsilon = 0.05$ we obtained

$$\tilde{\mathbb{Q}}^{\epsilon=0.05}_{\text{est.}} = \begin{pmatrix} -4.7907 & 0.9942 & 3.7965 & 0 \\ 1.9958 & -6.1869 & 0 & 4.1911 \\ 4.8004 & 0 & -5.8066 & 1.0062 \\ 0 & 5.1638 & 1.9886 & -7.1524 \end{pmatrix},$$

i.e., deviations of about one percent. For $\epsilon = 0.2$ we obtained

$$\tilde{\mathbb{Q}}^{\epsilon=0.2}_{\text{est.}} = \begin{pmatrix} -4.8259 & 0.9955 & 3.8304 & 0 \\ 2.0033 & -6.2026 & 0 & 4.1993 \\ 4.8555 & 0 & -5.8644 & 1.0089 \\ 0 & 5.2026 & 1.9936 & -7.1962 \end{pmatrix},$$

i.e., deviations of about two percent. In either case the main source of error seems to be statistical. The fact that parameter estimation converges as $T \to \infty$ is by itself not surprising, since the whole process is ergodic. The question is to what extent is the reduced process $(X_1(t), X_3(t))$ Markovian? To test this we re-evaluated the entries of the rate matrix conditional on the preceding state. This yielded changes of about one percent, which again, could be attributed to a lack of statistics.

**Example 5.2** Consider next Example 4.4 with $S_j = \{0,1\}$, $j = 1,\ldots,6$, and conditional rate matrices,

$$\mathbb{Q}^1 = \begin{pmatrix} -1 & 1 \\ 2 & -2 \end{pmatrix} \quad \mathbb{Q}^2 = \begin{pmatrix} -2 & 2 \\ 1 & -1 \end{pmatrix}$$

$$\mathbb{Q}^{3|1}_{\cdot|0} = \begin{pmatrix} -2 & 2 \\ 3 & -3 \end{pmatrix} \quad \mathbb{Q}^{3|1}_{\cdot|1} = \begin{pmatrix} -3 & 3 \\ 2 & -2 \end{pmatrix}$$

$$\mathbb{Q}^{4|2,3}_{\cdot|(0,0)} = \begin{pmatrix} -3 & 3 \\ 4 & -4 \end{pmatrix} \quad \mathbb{Q}^{4|2,3}_{\cdot|(1,0)} = \begin{pmatrix} -4 & 4 \\ 3 & -3 \end{pmatrix}$$

$$\mathbb{Q}^{4|2,3}_{\cdot|(0,1)} = \begin{pmatrix} -1 & 1 \\ 2 & -2 \end{pmatrix} \quad \mathbb{Q}^{4|2,3}_{\cdot|(1,1)} = \begin{pmatrix} -2 & 2 \\ 1 & -1 \end{pmatrix}$$

$$\mathbb{Q}^{5|3}_{\cdot|0} = \begin{pmatrix} -1 & 1 \\ 3 & -3 \end{pmatrix} \quad \mathbb{Q}^{5|3}_{\cdot|1} = \begin{pmatrix} -3 & 3 \\ 1 & -1 \end{pmatrix}$$

$$\mathbb{Q}^{6|4}_{\cdot|0} = \begin{pmatrix} -1 & 1 \\ 4 & -4 \end{pmatrix} \quad \mathbb{Q}^{6|4}_{\cdot|1} = \begin{pmatrix} -4 & 4 \\ 1 & -1 \end{pmatrix}.$$

The slow component $X_5$ is only influenced by the fast component $X_3$, whose conditional equilibrium distribution only

depends on its slow ancestor $X_1$:

$$\pi^{3|1}_{\cdot|0} = \frac{1}{5}\begin{pmatrix}3\\2\end{pmatrix} \qquad \pi^{3|1}_{\cdot|1} = \frac{1}{5}\begin{pmatrix}2\\3\end{pmatrix}.$$

As $\epsilon \to 0$ the conditional rate matrix associated with $X_5$ converges to

$$\tilde{Q}^{5|1}_{\cdot|0} = \pi^{3|1}_{0|0}Q^{5|3}_{\cdot|0} + \pi^{3|1}_{1|0}Q^{5|3}_{\cdot|1} = \frac{1}{5}\begin{pmatrix}-9 & 9\\11 & -11\end{pmatrix}$$

$$\tilde{Q}^{5|1}_{\cdot|1} = \pi^{3|1}_{0|1}Q^{5|3}_{\cdot|0} + \pi^{3|1}_{1|1}Q^{5|3}_{\cdot|1} = \frac{1}{5}\begin{pmatrix}-11 & 11\\9 & -9\end{pmatrix}.$$

To test the reduction procedure, we generated a trajectory of the full process with $10^7$ transitions and $\epsilon = 0.05$. We used maximum likelihood to estimate the rates associated with the reduced process that only consists of the slow components $X_1, X_2, X_5, X_6$. For example, we estimated the rates associated with $X_5$ transitioning from 0 to 1; one can estimate these rates assuming that they depend on the state of all other (slow) components. The estimation procedure gives

$$q^{5|1,2,6}_{0,1|(0,0,0)} = 1.78 \qquad q^{5|1,2,6}_{0,1|(1,0,0)} = 2.19$$
$$q^{5|1,2,6}_{0,1|(0,1,0)} = 1.77 \qquad q^{5|1,2,6}_{0,1|(1,1,0)} = 2.20$$
$$q^{5|1,2,6}_{0,1|(0,0,1)} = 1.77 \qquad q^{5|1,2,6}_{0,1|(1,0,1)} = 2.18$$
$$q^{5|1,2,6}_{0,1|(0,1,1)} = 1.80 \qquad q^{5|1,2,6}_{0,1|(1,1,1)} = 2.18.$$

These results are in very good agreement with our prediction that the rates in the left column be equal to $9/5 = 1.8$ and the rates in the right column to $11/5 = 2.2$, irrespectively of the values of $X_2$ and $X_6$.

Finally, we show in Table 1 the estimated values of $\tilde{q}^{5|1}_{0,1|0}$ and $\tilde{q}^{5|1}_{0,1|1}$ for various values of $\epsilon$. The case $\epsilon = 1$ means that all components evolve on comparable time scales, in which case the reduction procedure does not hold. This table confirms the intuitive feeling that the approximation gets reasonably accurate for $\epsilon$ of the order of 0.1. Note that the estimate of $\tilde{q}^{5|1}_{0,1|1}$ for $\epsilon = 0.025$ is less accurate than for $\epsilon = 0.05$; this is attributed to the fact that the smaller $\epsilon$, the less (relatively) probable it is to observe a transition in the slow variables, resulting in poor statistics.

## 6 Discussion

In this paper we have proved a theorem about dimension reduction of CTBNs in the limit of an infinite separation of scales between fast and slow components. We showed the implications of this theorem for constructing a reduced CTBN that captures the dynamics of the slow components without explicitly dealing with the fast ones.

Our results show that the elimination of fast components has a counter intuitive property. The typical intuition is that integrating out a variable introduces dependencies among its children. However, when eliminating fast components this intuition does not apply directly, and in some cases we end with a simpler CTBN than we started with. Thus, our reduction leads to further simplifications than one might expect from basic intuitions about Bayesian networks.

| $\epsilon$ | $\tilde{q}^{5|1}_{0,1|0}$ | $\tilde{q}^{5|1}_{0,1|1}$ |
|---|---|---|
| 1 | 1.638 | 1.875 |
| 0.5 | 1.700 | 1.992 |
| 0.25 | 1.742 | 2.077 |
| 0.1 | 1.766 | 2.145 |
| 0.05 | 1.782 | 2.189 |
| 0.025 | 1.796 | 2.174 |
| $\to 0$ | 1.800 | 2.200 |

Table 1: Estimated values of entries of the (effective) conditional rate matrix $\tilde{Q}^{5|1}$ for various values of $\epsilon$. The maximum likelihood estimation was applied to trajectories (of the full system) with $10^7$ transitions.

In practice, one seldom encounters systems in which a small parameter $\epsilon$ is explicitly given. In many applications there exists a range of characteristic rates, and one has to verify to what extent the dimension reduction is a good approximation. Since equilibration is exponentially fast, dimensional reduction is expected to be a good approximation when the equilibration rates associated with a subset of components are larger, by at least an order of magnitude, than the equilibration rates associated with the other components.

To put the results we introduce here to use we need to develop them into concrete approximation algorithms. The appeal of such results is that they give us a strategy to use separation of scales to reason about the system at different levels of time granularity. For reasoning about coarse time scales, our results allow to reduce the system to examine only the slow components. To reason about fine time scales we can then assume that most of the slow components are fixed, and then reason about the dynamics of the fast components. Clearly this intuition can be extended to a hierarchy of time scales.

Given a CTBN, we can assess the characteristic equilibration rate of each conditional Q-matrix by computing the absolute value of its second largest eigenvalue. There are, however, multiple ways of using these values to separate the system into an approximate hierarchy of scales. Another issue deals with evidence. Clearly, once we find a reduced CTBN we can incorporate evidence and reason about the posterior probability of slow components and consequently fast components. However, it is also fairly clear that the frequency of observations and the time scale of the observed variables can make important impact on the approximation.

The results we presented here provide solid foundations for introducing scale-based approximation in real applications. Clearly, these initial results are only the first step in the development of promising approximate inference procedures.

## Acknowledgments

Nir Friedman was supported in part by grants from the Israel Science Foundation (ISF) and from the Binational US-Israel Science Foundation (BSF). Raz Kupferman was supported in part by a grant from the Israel Science Foundation (ISF).

## A  Proof sketch of Theorem 3.1

Using the partition of the $\mathbb{Q}$-matrix as the sum of fast and slow components, the master equation (1) takes the form

$$\frac{d}{dt}p - \frac{1}{\epsilon}(\mathbb{Q}^{\text{fast}})^T p = (\mathbb{Q}^{\text{slow}})^T p.$$

If we treat the right hand side as an inhomogeneous term, this equation can be integrated, resulting in an integral equation,

$$p(t) = e^{\frac{t}{\epsilon}(\mathbb{Q}^{\text{fast}})^T}p(0) + \int_0^t e^{\frac{(t-s)}{\epsilon}(\mathbb{Q}^{\text{fast}})^T}(\mathbb{Q}^{\text{slow}})^T p(s)\,ds.$$

The one-parameter semigroup of operators $\exp(t\mathbb{Q}^{\text{fast}})$ is the solution operator of the master equation derived from the fast dynamics, with the slow components held fixed.

Assumption 3.1 implies that $\exp(t\mathbb{Q}^{\text{fast}})$ converges exponentially fast, as $t \to \infty$, to an operator $\mathbb{G}$, which is the orthogonal projection onto the subspace of distributions that are invariant under $\mathbb{Q}^{\text{fast}}$. The projection $\mathbb{G}$ has entries

$$g_{a,b} = \pi_{\text{Fast}(b)|\text{Slow}(b)}^{I_{\text{fast}}|I_{\text{slow}}}\,\delta_{\text{Slow}(a),\text{Slow}(b)}.$$

As $\epsilon \to 0$, the distribution $p(t)$ tends to the solution of the limiting equation,

$$p(t) = \mathbb{G}^T p(0) + \int_0^t \mathbb{G}^T(\mathbb{Q}^{\text{slow}})^T p(s)\,ds,$$

which is equivalent to the differential equation,

$$\frac{d}{dt}p = \mathbb{G}^T(\mathbb{Q}^{\text{slow}})^T p. \qquad (10)$$

Note that according to the limiting equation $p(0)$ is in the range of the projection $\mathbb{G}^T$; if this assumption does not hold, the actual limit of $p(t)$ deviates from the solution of the limiting equation only in a short time interval after the initial time (a "boundary layer").

The range of $\mathbb{G}^T$, which consists of distributions of the form

$$p_a = \pi_{\text{Fast}(a)|\text{Slow}(a)}^{I_{\text{fast}}|I_{\text{slow}}}\tilde{p}_{\text{Slow}(a)},$$

where $\tilde{p}$ is the marginal distribution over $S_{\text{slow}}$, is invariant under Equation (10). Substituting this product into Equation (10) and summing over all $\text{Fast}(a)$, we get an equation for the marginal distribution,

$$\frac{d}{dt}\tilde{p}_\alpha = \sum_{\beta \in S_{\text{slow}}}\left(\sum_{\zeta \in S_{\text{fast}}} q^{\text{slow}}_{(\beta,\zeta),(\alpha,\zeta)}\pi_{\zeta|\beta}^{I_{\text{fast}}|I_{\text{slow}}}\right)\tilde{p}_\beta.$$

Comparing with (8), the expression in the brackets is identified as $\tilde{q}_{\alpha,\beta}$, i.e., $\tilde{p}$ satisfies a master equation with rate matrix $\tilde{\mathbb{Q}}$.